\documentclass[letterpaper, 10 pt, journal, twoside]{IEEEtran}
\usepackage{amsmath,amsfonts}
\usepackage{algorithmic}
\usepackage{algorithm}
\usepackage{array}
\usepackage[caption=false,font=normalsize,labelfont=sf,textfont=sf]{subfig}
\usepackage{textcomp}
\usepackage{stfloats}
\usepackage{verbatim}
\usepackage{graphicx}
\usepackage{booktabs}
\usepackage{xurl}
\usepackage{cite}
\usepackage{supertabular}
\usepackage{makecell}
\begin{document}

\title{ForaNav: Insect-inspired Online Target-oriented Navigation for MAVs in Tree Plantations}

\author{Weijie Kuang$^{1}$, Hann Woei Ho$^{1,*}$,~\IEEEmembership{Member,~IEEE}, Ye Zhou$^{1}$ and Shahrel Azmin Suandi$^{2}$,~\IEEEmembership{Senior Member,~IEEE}
\thanks{The authors would like to thank Malaysian Ministry of Higher Education (MOHE) for providing the Fundamental Research Grant Scheme (FRGS) (Grant number: FRGS/1/2024/TK04/USM/02/3) for conducting this research. The first author would like to thank the Chinese Scholarship Council (CSC) for the Ph.D. financial support (Project number: 202106830031). \textit{($^{*}$Corresponding author: Hann Woei Ho)}
}
\thanks{$^{1}$Weijie Kuang, Hann Woei Ho, and Ye Zhou are with School of Aerospace Engineering, Engineering Campus, Universiti Sains Malaysia, 14300 Nibong Tebal, Pulau Pinang, Malaysia (email: wjkuang@student.usm.my; aehannwoei@usm.my; zhouye@usm.my).
}
\thanks{$^{2}$Shahrel Azmin Suandi is with School of Electrical and Electronic Engineering, Engineering Campus, Universiti Sains Malaysia, 14300 Nibong Tebal, Pulau Pinang, Malaysia (email: shahrel@usm.my).
}

}

\markboth{This paper has been submitted to IEEE Robotics and Automation Letters.}%
{Kuang \MakeLowercase{\textit{et al.}}: ForaNav: Insect-inspired Online Target-oriented Navigation for MAVs in Tree Plantations}

\maketitle

\begin{abstract}
Autonomous Micro Air Vehicles (MAVs) are becoming essential in precision agriculture to enhance efficiency and reduce labor costs through targeted, real-time operations. However, existing unmanned systems often rely on GPS-based navigation, which is prone to inaccuracies in rural areas and limits flight paths to predefined routes, resulting in operational inefficiencies. To address these challenges, this paper presents ForaNav, an insect-inspired navigation strategy for autonomous navigation in plantations. The proposed method employs an enhanced Histogram of Oriented Gradient (HOG)-based tree detection approach, integrating hue-saturation histograms and global HOG feature variance with hierarchical HOG extraction to distinguish oil palm trees from visually similar objects. Inspired by insect foraging behavior, the MAV dynamically adjusts its path based on detected trees and employs a recovery mechanism to stay on course if a target is temporarily lost. We demonstrate that our detection method generalizes well to different tree types while maintaining lower CPU usage, lower temperature, and higher FPS than lightweight deep learning models, making it well-suited for real-time applications. Flight test results across diverse real-world scenarios show that the MAV successfully detects and approaches all trees without prior tree location, validating its effectiveness for agricultural automation.
\end{abstract}


\section{Introduction}
\IEEEPARstart{P}{recision} farming is essential for improving agricultural yields while minimizing environmental risks, aligning with the Sustainable Development Goals. Agricultural tasks such as crop monitoring and targeted interventions have traditionally demanded substantial labor and resources. Recently, Micro Air Vehicles (MAVs) have emerged as a highly promising solution to automate these tasks.

\begin{figure}[!t]
\centering
  \includegraphics[width=3in]{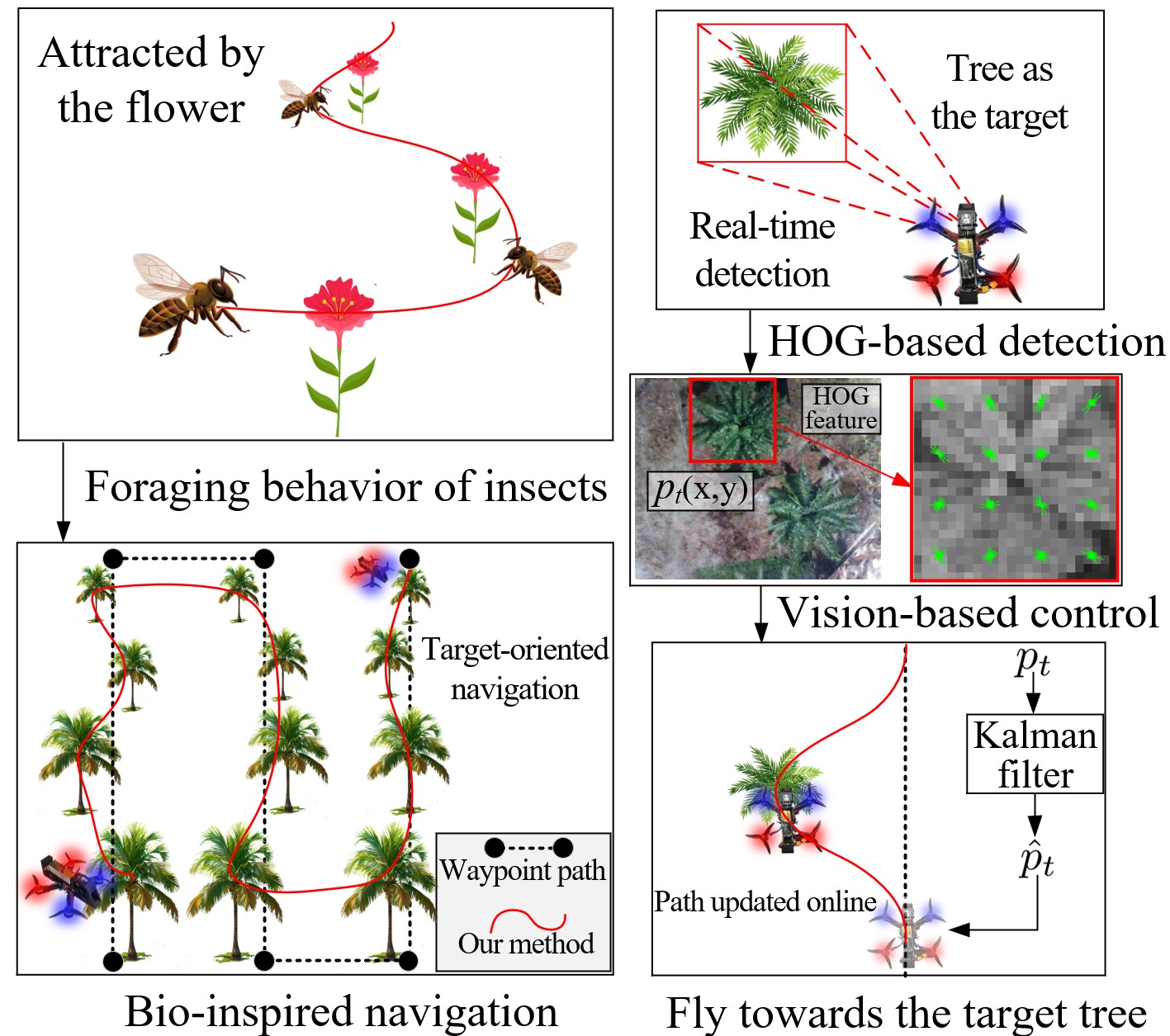}
  \caption{Schematic of insect-inspired egocentric visual navigation. Inspired by the foraging behavior of insects, the MAV navigates with targeting oil palm trees. The MAV detects trees in real time using an enhanced HOG-based method. Their image coordinates $p_{t}$ are tracked with a Kalman filter to estimate $\hat{p}_{t}$ which is used to update MAV trajectory online to approach each target.}
  \label{Navigationschematic}
\end{figure}

Current MAV navigation in agriculture relies on GPS waypoints at the start and end of each tree column, forming predefined rectangular flight routes (represented by dotted lines on the bottom left of Fig. \ref{Navigationschematic}) \cite{wei2024precise}. While this provides structured coverage, it does not account for individual tree positions. RTK GPS can offer accurate coordinates, but collecting this data requires manual effort under ideal conditions. Moreover, plantations are dynamic, with tree growth, fall, and disease altering tree positions over time. Thus, the \textit{key challenge} of this common approach is its potential for inefficient coverage and missed targets. To address this, real-time tree detection is essential for MAVs to dynamically adjust flight paths based on detected tree information. This enables more targeted interventions in precision agriculture, including canopy size assessment and health evaluation \cite{kuang2024comprehensive}.

Implementing real-time tree detection for MAV navigation is challenging due to strict size and weight constraints, which limit payload capacity and onboard processing power. Unlike ground vehicles, MAVs cannot support advanced algorithms that rely on large sensors or heavy computation. Therefore, efficient navigation solutions require small and lightweight sensors while maintaining accuracy and reliability \cite{chen2024fastnav}.

An effective solution can be drawn from nature, specifically from the foraging behavior of insects, where egocentric navigation plays a key role \cite{basu2024neural}. Insects use their current view to determine movement direction, enabling efficient foraging. This egocentric representation simplifies computation, as spatial changes map directly onto shifts in view, eliminating the need for complex transformations. 
Meanwhile, many insects also rely on the ability of odors localization by chemical sensing to detect and locate the source of potential food sources \cite{anderson2020bio}. These behaviors demonstrate how insects plan their routes based on resource locations. This enables them to gather nectar quickly through direct and efficient navigation. Similarly, this strategy can inspire MAV navigation strategy to optimize flight paths for precision agricultural tasks, such as targeted orchard monitoring.

Thus, this study aims to develop a solution that enables MAVs to navigate over plantations with real-time tree tracking. Inspired by the foraging behavior of insects attracted by food, we propose ForaNav, an online target-oriented navigation method.
We chose oil palm trees as the detection target because of their dense canopy structure and complex plantation layouts. Their economic importance also makes them an ideal model for solutions applicable to other crops.
The \textit{main contributions} of this paper are:
\begin{itemize}
\item a real-time palm tree detection algorithm using an enhanced hierarchical HOG method was developed, which is capable of distinguishing the targets from other visually similar objects in the environment and generalizing to different tree types,
\item a navigation strategy inspired by insect foraging behavior was designed, specifically tailored for real-world plantation deployment and tested across various challenging layouts—aligned, irregular, and clustered trees—demonstrating its effectiveness in diverse environments, and
\item a robust recovery mechanism in navigation inspired by the view memory of insects was created to ensure MAVs remain on course despite temporary target loss.
\end{itemize}

The remainder of this paper is organized as follows: Section \ref{sec:relatedworks} presents a review of the relevant literature. In Section \ref{sec:method}, the proposed method is detailed. Section \ref{sec:results} outlines the flight experiments and results discussion, and finally, the conclusion is drawn in Section \ref{sec:conclusion}.

\section{Related works}
\label{sec:relatedworks}
Accurate tree detection is essential for precision plantation management, enabling effective monitoring and decision-making. Deep learning-based methods have gained prominence due to their exceptional accuracy. For instance, a model integrating a refined pyramid feature module and a hybrid class-balanced loss module with Faster RCNN achieved an impressive F1-score of 99.04\% in oil palm detection \cite{zheng2021growing}. Similarly, Yolo-V5 outperformed other CNN architectures, including SSD300, Yolo-V3, and Yolo-V4, for palm tree detection \cite{jintasuttisak2022deep}. More recently, an enhanced RetinaNet model was designed to improve the detection of young and mature oil palm trees in complex environments, which showed strong generalization and accuracy \cite{lee2024oil}. Despite the high accuracy of these deep learning-based techniques, their computational demands pose challenges for real-time applications.

In contrast, several oil palm tree detection methods bypass deep learning. One approach used scale-invariant feature transform (SIFT) to extract keypoints from MAV images. A pre-trained extreme learning machine classifier then analyzed these keypoints to differentiate palm from non-palm features \cite{malek2014efficient}.
In another study, an image thresholding method was developed to detect oil palm trees \cite{putra2022oil}. In this method, images were segmented to isolate trees from the background using morphological operations and the Otsu method. Tree detection was then performed through template matching on the segmented images.
Additionally, another approach combined Histogram of Oriented Gradient (HOG) and Support Vector Machine (SVM) to detect oil palm trees \cite{wang2019automatic}. In this method, HOG was applied as a feature descriptor, and the extracted features were classified by an SVM model, similar to the approach used for crater detection on the lunar \cite{wang2024small}. Besides using HOG-SVM to detect target directly, it can also be used to extract regions of interest, which can then be combined with deep learning methods for the object detection, such as using ResNet-101 for traffic object classification \cite{shirpour2023traffic}. 

These previous studies used high-altitude imagery, i.e., over 100 meters, to capture more trees, resulting in lower resolution for individual trees. Although these methods performed well for tree detection, they are less suitable for low-altitude MAV missions. In these missions, images typically contain fewer trees and a larger proportion of background clutter, which complicates detection. Additionally, the feasibility of running these deep learning models in real-time on MAVs remains uncertain, as they have not yet been tested in onboard tasks.

In addition to detection, navigation is another crucial aspect of autonomous tasks of MAVs in agriculture. Several vehicle navigation studies take nature as an excellent source of inspiration for efficient solutions \cite{de2022insect}. For example, AntBot mimicked the remarkable navigation skill of desert ants by using optical sensors to process light polarization and optic flow \cite{dupeyroux2019antbot}. These sensors allow AntBot to estimate its orientation and the distance traveled to navigate in a GPS-denied outdoor environment.
Inspired by the plume tracking behavior of the moth, a study developed a bio-hybrid flight system capable of autonomously localizing odors \cite{anderson2020bio}.
Inspired by the view memory of insects, another study combined odometry and visual homing for tiny robots to visually navigate to the stored locations \cite{van2024visual}.
Such vision-based navigation allows robots to estimate their position with limited computing power and visibility, enabling efficient movement in cluttered spaces \cite{gummadi2024fed,li2025fact}.
These investigations demonstrate that tiny insects contain enormous potential and inspiration.

\begin{figure}[!t]
\centering
  \includegraphics[width=3.5in]{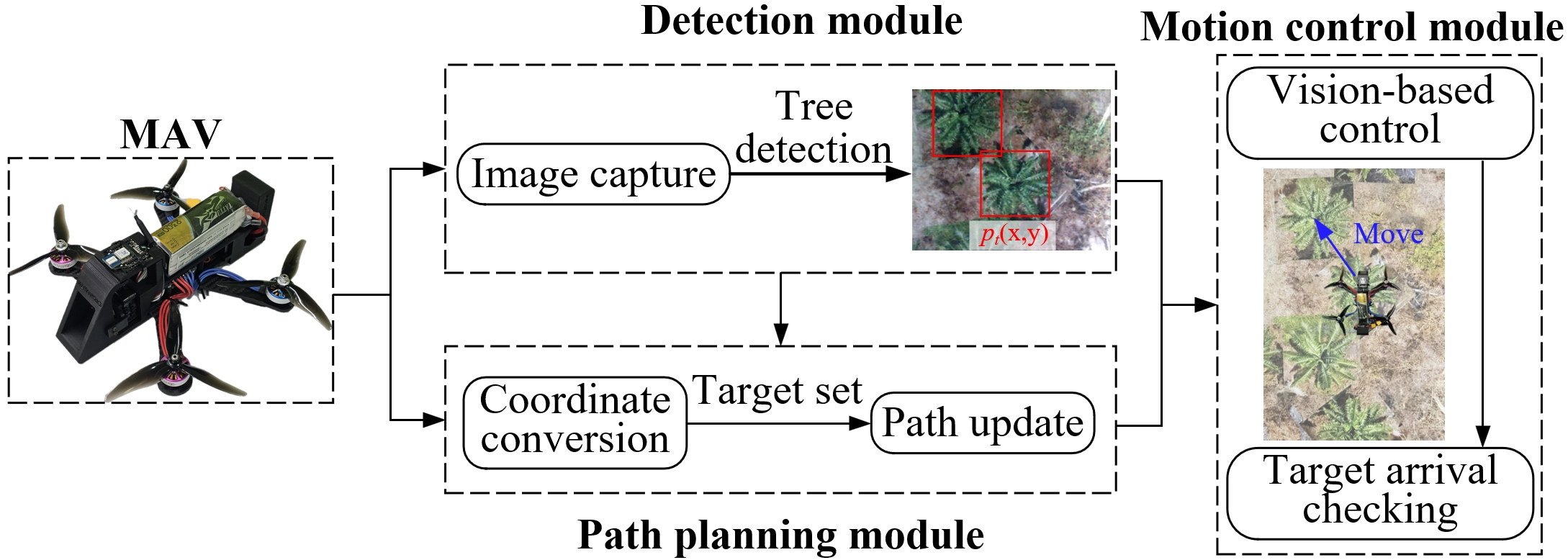}
  \caption{Overview of the detection and navigation frameworks. The onboard camera system captures images and detects trees in real time, using their image coordinates to update the MAV trajectory via the path planning module and track targets through the motion control module.}
  \label{navigation system}
\end{figure}

Therefore, this paper proposes an insect-inspired navigation strategy for MAVs. It is designed to perform tree detection and low-altitude plantation management tasks. The tree detection method integrates hue and saturation comparison with the variance of global HOG features. It employs hierarchical HOG feature extraction to distinguish color-similar and structure-similar objects from oil palm trees in cluttered environments. Meanwhile, the navigation strategy, inspired by the foraging behavior of insects, is implemented. It includes a recovery mechanism that allows MAVs to approach the detected trees without deviating from the route. The developed autonomous system is illustrated in Fig. \ref{navigation system}.

\section{Methodology}
\label{sec:method} 
This section presents the navigation framework, ForaNav, composed of two main parts: tree detection and insect-inspired navigation. During the flight, the tree detection is performed using the onboard camera. Once detection occurs, the MAV navigates toward the target using the insect-inspired strategy. 

\subsection{Oil palm tree detection}
To achieve real-time oil palm tree detection on MAVs, which is resource-restricted, we employed the concept from prior research \cite{wang2019automatic}. Specifically, it extracts star-shaped features of oil palm trees using the HOG algorithm and classifies them with an SVM model. However, two major problems appear when detection is conducted in images taken at low altitudes in cluttered environments. The first problem is distinguishing oil palm trees from other palm plants that exhibit similar star-shaped HOG features in their pinnate leaves. The second problem is that similar palm tree species also have star-shaped HOG features. Other background objects, like roofs and road intersections, also share similar characteristics.

\begin{figure}[!t]
\centering
  \includegraphics[width=3.5in]{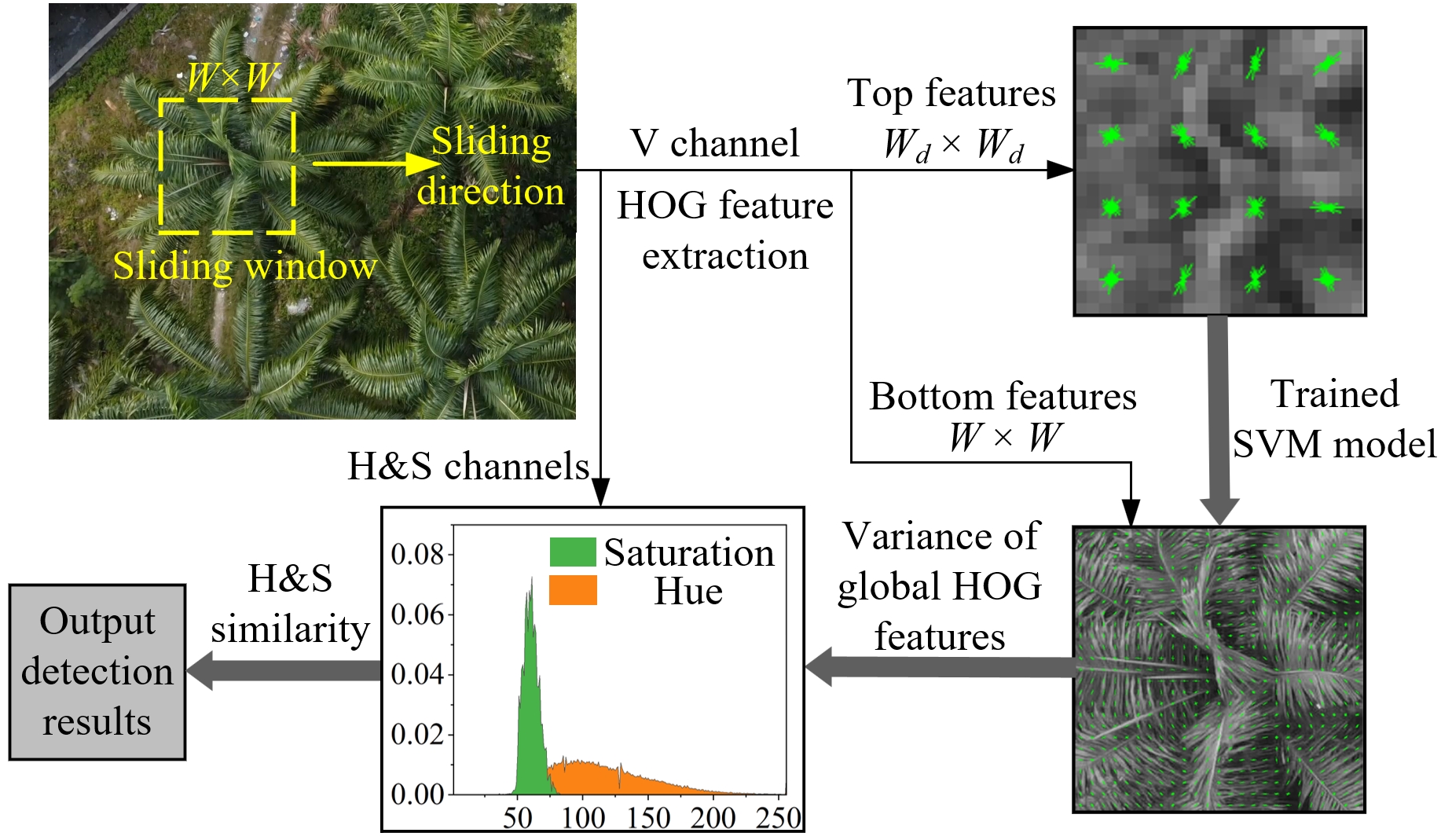}
  \caption{Workflow diagram of the proposed oil palm tree detection method. The detection window is classified as an oil palm tree based on SVM classification, variance calculations, and hue and saturation comparisons.}
  \label{Flowchart}
\end{figure}

An improved method is proposed to address these issues, with its workflow shown in Fig. \ref{Flowchart}. The sliding window of size $W$ scans the captured image with the step of one-third $W$ in both the x and y axes. The RGB image of the sliding window is transformed into the HSV format, containing hue, saturation, and value channels. The HOG algorithm effectively extracts the edge and shape features of oil palm trees in the value channel. Specifically, the HOG method divides an image into cells, calculates the gradient direction and magnitude for each pixel, and forms HOG orientations. In the HOG method \cite{dalal2005histograms}, the gradient is calculated by applying a convolution calculation, typically with a kernel of [-1, 0, 1], to determine changes in intensity along the x and y axes:
\begin{equation}
G_{x}(x,y) = I(x+1,y)-I(x-1,y),
\end{equation}
\begin{equation}
G_{y}(x,y) = I(x,y+1)-I(x,y-1).
\end{equation}
Here, $G_{x}$ and $G_{y}$ are the gradients in the x and y directions, respectively, and \(I(x, y)\) represents the pixel value. Then the magnitude $G$ and the direction \(\theta\) are calculated as follows:
\begin{equation}
G = \sqrt{G_{x}^2+G_{y}^2},
\end{equation}
\begin{equation}
\theta = arctan\frac{G_{y}}{G_{x}}.
\end{equation}

The calculation of HOGs depends on key parameters: cell size $C$, block size $B$, block stride $S$, histogram bins $n$, and window size $W_{d}$. Smaller cells and blocks capture finer details, while longer strides improve speed at the cost of accuracy. And more histogram bins provide finer orientation data.

\begin{figure}[!t]
    \centering
    \includegraphics[width=0.4\textwidth]{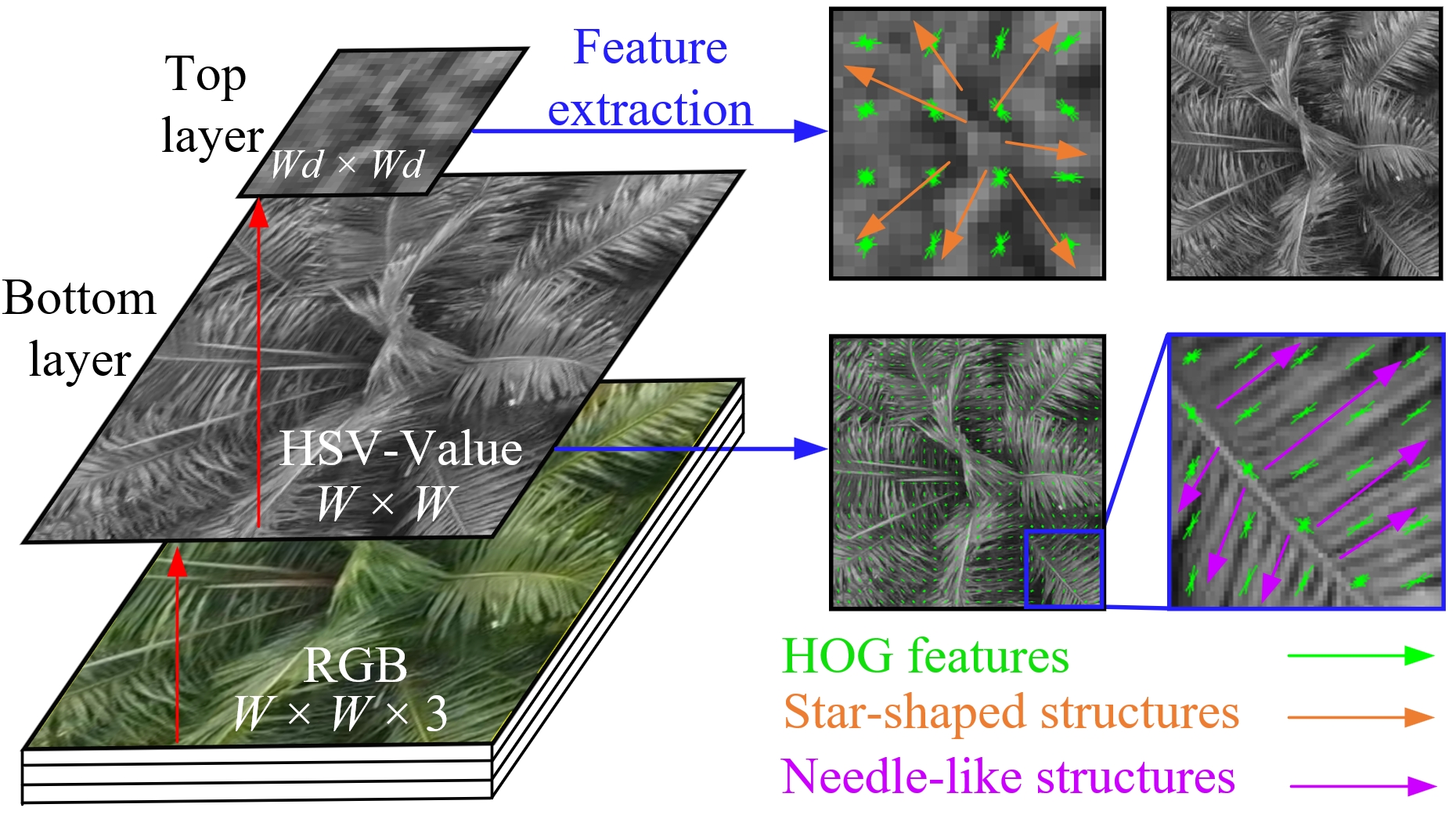}
    \caption{Schematic diagram of the hierarchical HOG feature extraction of oil palm trees. The green arrows represent the HOG features across different layers. The orange arrows depict the star-shaped HOG features of pinnate leaves. The purple arrows show the needle-like structures of leaflets.}
    \label{Pyrammid}
\end{figure}

To address the first problem, the needle-like leaflets of oil palms can differentiate them from similar species.
Therefore, we propose a hierarchical HOG feature extraction of oil palm trees, where the bottom layer utilizes images of $W$ × $W$ pixels to get needle-like features of leaflets. And the top layer uses down-sampled images of $W_{d}$ × $W_{d}$ pixels to get star-shaped features of pinnate leaves. The schematic diagram of the hierarchical extraction method is depicted in Fig. \ref{Pyrammid}.

Then, the global HOG features of the bottom layer are used to calculate the variance \(Var_{hog}\), which is the average of the squared deviations from the mean:
\begin{equation}
Var_{hog} = \frac {1}{n}\sum _{i=1}^{n}(h_{i} - \bar {h_{i}})^{2},
\end{equation}
where \(h_{i}\) represents the $i$-th number in HOG, \(\bar {h_{i}}\) is the mean of HOG, and \(n\) is the bin number in HOG. The variance of the HOG features of oil palm trees should be larger due to the fine texture of needle-like leaflets. After extracting HOG features from the top layer, an SVM classifier needs to be trained to identify the star-shaped structures.

To solve the second problem, the difference in the hue and saturation channels between oil palm trees and other background objects is considered. Specifically, the hue and saturation histograms of an image are generated for comparison with those of a template oil palm tree image. The Pearson correlation coefficient \cite{pearson1896vii} is applied to evaluate linear relationships between hue or saturation histograms, providing information on how their values change in tandem:
\begin{equation}
H_{corrs} = \frac{\sum_{i} (H_{\text{ref}}[i] - \bar{H}_{\text{ref}}) (H_{\text{cur}}[i] - \bar{H}_{\text{cur}})}{
\sigma_{\text{ref}} \cdot \sigma_{\text{cur}} \cdot 255},
\end{equation}
where \(H_\text{ref}[i]\) and \(H_\text{cur}[i]\) are the elements of the template and current histograms, respectively. \(\bar{H}_\text{ref}\) and \(\bar{H}_\text{cur}\) are their means, while \(\sigma_\text{ref}\) and \(\sigma_\text{cur}\) are their standard deviations. The detected trees serve as key inputs for our target-oriented navigation strategy, which dynamically adjusts the MAV trajectory. 

\subsection{Insect-inspired navigation strategy in plantations}

Our navigation strategy draws inspiration from insect foraging behavior \cite{freire2025salt}, particularly the reliance on local visual cues and their body-centered frame of reference for position estimation and efficient movement. For instance, bees and ants utilize egocentric visual navigation to locate and approach food sources, continuously refining their trajectories through sensory feedback \cite{basu2024neural}. Similarly, the MAV detects trees as targets in real time using its onboard camera and dynamically adjusts its path to minimize its distance to targets. Furthermore, our recovery mechanism is inspired by view memory in insect navigation, where it estimates and temporarily stores tree positions. This enables the MAV to recover from temporary occlusions, maintaining trajectory robustness in cluttered environments.

The target-oriented navigation strategy is outlined in Algorithm \ref{alg1}. The input of this algorithm is the predetermined flight path \(F\), defined only by the starting and ending waypoints of each planting column. Initially, the MAV follows \(F\). Upon detecting a tree with image coordinates $p_{t}$, a Kalman filter predicts and updates the tree image coordinates $\hat{p}_{t}$ for stable tracking. The MAV adjusts its movement to minimize the offset $e_{p}$ between $\hat{p}_{t}$ and the image center to approach the target tree $t^*$. The tree is considered reached only when it is nearly centered in the image, determined with a threshold $\tau_{p_{th}}$. 

To handle temporary target loss, a recovery mechanism inspired by insect view memory \cite{cartwright1982honey} is implemented. Each detected tree position $s_t$ is estimated using $\hat{p}_{t}$ and the MAV position $s_{\text{MAV}}$ and stored as $T_d$. If the deviation $e_{d}$ between $s_{\text{MAV}}$ and $s_t$ exceeds the threshold $\delta_{d_{th}}$, the MAV re-approaches $s_t$ until the tree is detected again. Once $p_{t}$ reacquired, visual guidance resumes by minimizing $e_{p}$. 

While this mechanism enhances robustness, its effectiveness may be limited in densely planted areas where overlapping canopies complicate tree differentiation. Excessive canopy overlap can distort HOG features, leading to inaccurate detection. Nevertheless, the proposed navigation method enables the MAV to autonomously traverse plantations, dynamically adjusting its path based on real-time tree detection. The Kalman filter ensures stable tracking, while the recovery mechanism improves reliability, reducing the risk of missed trees.

\begin{algorithm}[H]
\caption{ Insect-inspired autonomous navigation}
\begin{algorithmic}[1]
\STATE {\textbf{Input:}} Flight path  \(F\) generated by waypoints

\STATE {\textbf{Output:}} Updated \(F\) using tree detection results

\STATE {Set flight path \(F\)}

\STATE Initialize detected tree set: $T_{d}$ 

\STATE {\textbf{while}} Flight path \(F\) is not completed \textbf{do}

\STATE \hspace{0.5cm} Fly along flight path \(F\)

\STATE \hspace{0.5cm} Detected tree image coordinates: $p_{t}$ 

\STATE \hspace{0.5cm} {\textbf{if}} $p_{t} \neq \emptyset$ {\textbf{then}}

\STATE \hspace{1cm} Get MAV position $s_{\text{MAV}}$

\STATE \hspace{1cm} Estimate tree position $s_{t}$

\STATE \hspace{1cm} $T_{d} \leftarrow T_{d}\cup$\{$(p_{t},s_{t})$\}  

\STATE \hspace{0.5cm} {\textbf{end if}}

\STATE \hspace{0.5cm} {\textbf{while}} MAV has not reached the target tree $t^*$ {\textbf{do}}

\STATE \hspace{1.0cm} $ \hat{p}_{t} \leftarrow$ KalmanFilter($p_{t}$)

\STATE \hspace{1.0cm} Calculate the offset $e_{p}$ by $ \hat{p}_{t}$

\STATE \hspace{1.0cm} Approach to the tree by minimizing $e_{p}$

\STATE \hspace{1cm} {\textbf{if}} $e_{p} < \tau_{p_{th}}$

\STATE \hspace{1.5cm} {\textbf{break}}

\STATE \hspace{1cm} {\textbf{end if}}

\STATE \hspace{1.0cm} Calculate the deviation $e_{d}$ by $s_{\text{MAV}}, s_{t}$

\STATE \hspace{1cm} {\textbf{while}} $e_{d} > \delta_{d_{th}}$

\STATE \hspace{1.5cm} approach to $s_{t}$

\STATE \hspace{1cm} {\textbf{end while}} 

\STATE \hspace{0.5cm} {\textbf{end while}} 

\STATE \hspace{0.5cm} Remove tree:  $T_{d} \leftarrow T_{d}\backslash$\{$t^*$\} 

\STATE {\textbf{end while}}
\end{algorithmic}
\label{alg1}
\end{algorithm}

\section{Flight experiments and results}
\label{sec:results}
This section presents the flight tests conducted to validate the proposed methods. We outline the experimental setup, including MAV specifications and tree detection parameters. In addition to palm trees, we evaluate our detection method on two other tree types to assess its generalization. We also compare its real-time performance with other lightweight deep learning models. The results from nine flight tests across three different plantation layouts are analyzed, demonstrating the effectiveness of our navigation strategy in diverse scenarios.

\subsection{Experiment setup}
The experiments use a custom-built MAV which is equipped with a Crazyflie Bolt 1.1 flight controller, a Flowdeck V2 deck, a Loco Positioning deck, and a JeVois-A33 camera. As shown in Fig. \ref{navigation system}, the camera processes images to detect oil palm trees and sends their image positions to the flight controller. The flight controller consists of a path planning module, which updates the flight path, and a motion control module, which uses the image positions to guide the MAV toward the trees.


For oil palm tree detection, the window size $W$ of 300 pixels is used, and the image is downscaled to a window size $W_{d}$ of 24 pixels to extract star-shaped features. The block size $B$ is 3 cells, block stride $S$ is 1, and cell size $C$ is 6 pixels to capture structural details and minimize noise. The histogram bin number $C$ is set to 9 to represent the branches in various orientations. To train the SVM model, 800 images captured over a plantation are selected as the training set. To evaluate the trained model, a test set is built including 120 images as positive samples, and 120 images as negative samples, which do not contain complete trees.

To assess the generalization of our method, we applied it to detecting two other commercial tree species, citrus and olive trees, using datasets from \cite{Hnida2024Olive, niedz2025image}. Additionally, we compared the performance of the proposed HOG-based method with two lightweight deep learning models: MobileNetV2-SSD, designed manually, and EfficientDet, optimized through NAS. These models were chosen for their balance of speed and accuracy, lightweight structure, and suitability for resource-constrained systems \cite{ khan2025ripscout}.


\begin{figure}[!t]
    \centering
    \includegraphics[width=3.0in]{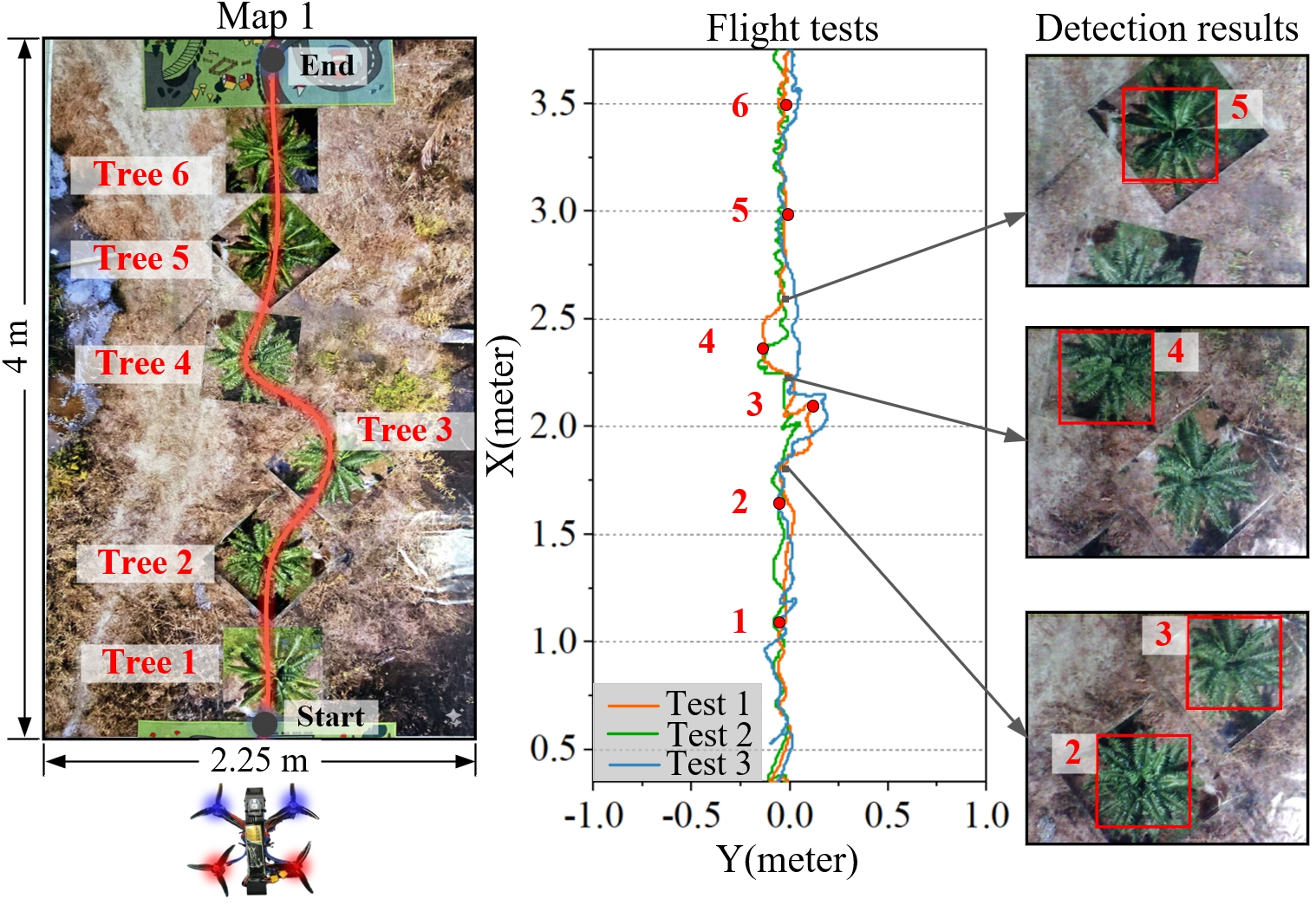}
    \caption{Flight test results in Map 1. Flight trajectories for Tests 1-3 are plotted in the center. Sample detection results from Test 1 are showed on the right.}
    \label{FlightResults1}
\end{figure}

\begin{figure}[!t]
    \centering
    \includegraphics[width=3.5in]{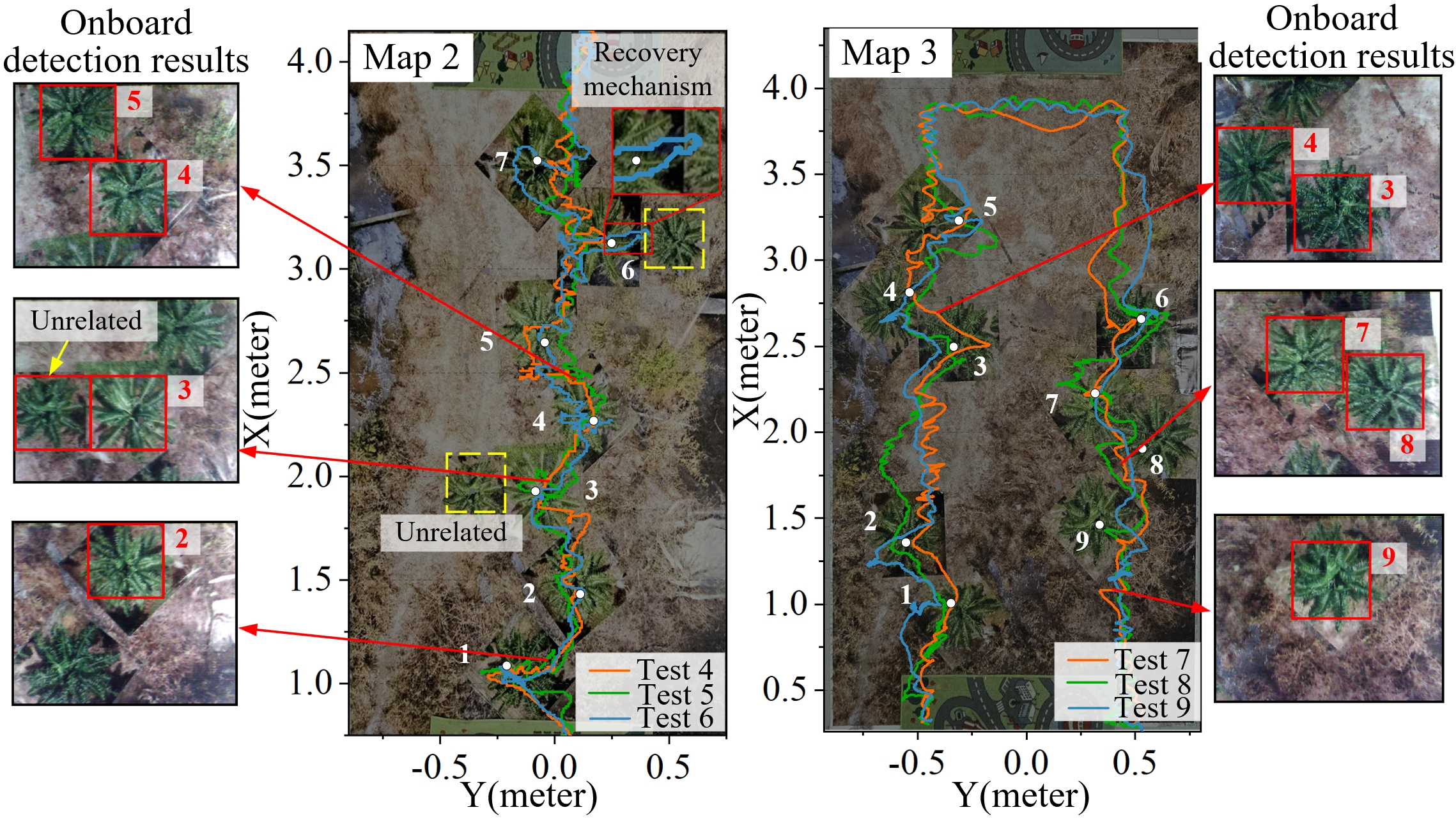}
    \caption{Flight test results in Map 2 and Map 3. Flight trajectories for Tests 4-9 are plotted on the maps. Sample detection results from Test 4 and Test 7 are showed on each side.}
    \label{FlightResults2}
\end{figure}

To validate the proposed solutions for autonomous navigation in plantation environments, we conducted nine tests across three different maps. Tests 1, 2, and 3 were performed on Map 1 (Fig. \ref{FlightResults1}), which features six trees arranged in a planting column, with Tree 3 slightly offset. Tests 4, 5, and 6 were conducted on Map 2, while Tests 7, 8, and 9 were performed on Map 3, both shown in Fig. \ref{FlightResults2}. Map 2 introduces a more challenging layout with seven trees positioned on both sides of one planting column, in which two of them are non-target trees. Map 3, designed to reflect complex real-world plantations, features an irregular distribution of nine trees across two planting columns, with some forming clusters.

The MAV uses orthogonal distances between detected trees and the flight path to check if they fall within the planting column. Trees that do not meet this criterion are disregarded. The planting distance is set to at least the crown size of a mature oil palm tree. Therefore, we set the threshold for the orthogonal distance at 1.5 times the crown size. 
All flight tests were conducted indoors to enable validation using an Ultra Wideband positioning system, which provides accuracy comparable to RTK GPS. While indoor conditions can be less favorable than outdoor environments—potentially affecting onboard detection due to lower lighting and minor turbulence—the controlled setting ensures consistency and repeatability, allowing for a reliable performance evaluation.

\subsection{Results and discussion}
\begin{figure}[!t]
    \centering
    \includegraphics[width=3.5in]{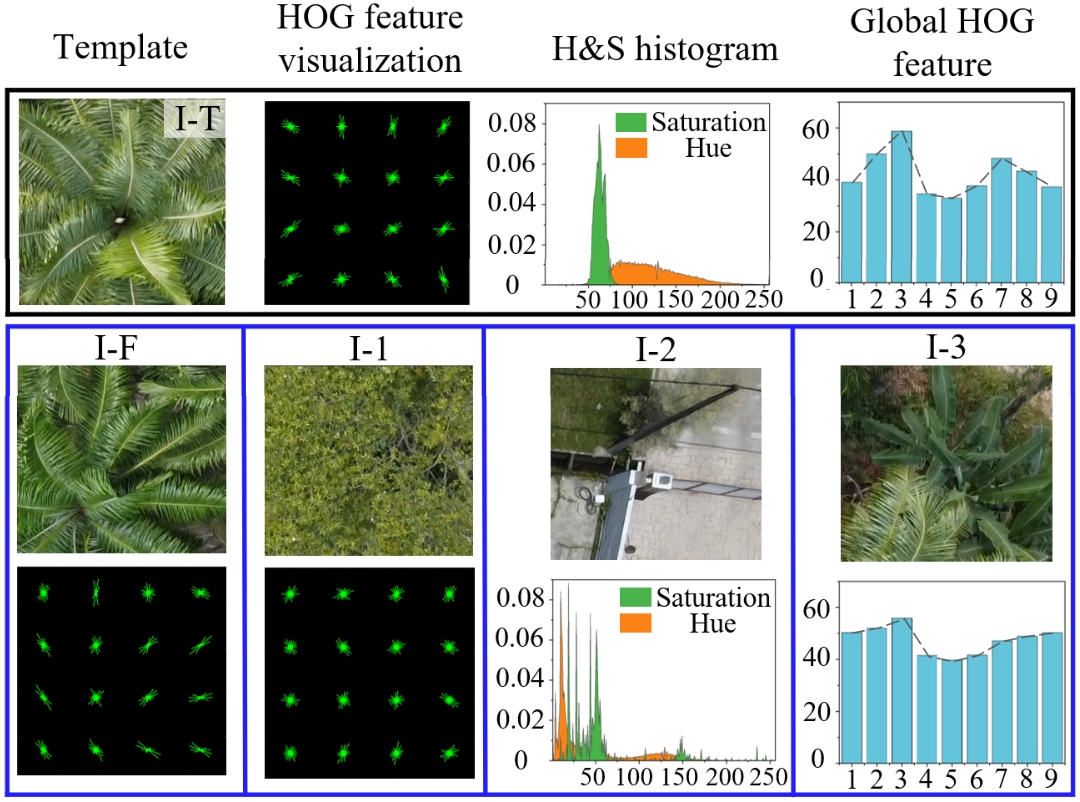}
    \caption{Comparison of extracted features in oil palm tree detection. The template image is compared against three background images (I-1, I-2, and I-3) using HOG feature visualization, H\&S histograms, and global HOG feature variance. Image I-F shows clustered palm trees.}
    \label{HSfunction}
\end{figure}

To evaluate the effectiveness of key design features in our detection method, we compare a template image (I-T) with three test images (I-1, I-2, and I-3). The comparison is made using HOG visualizations, H\&S histograms, and global HOG feature variance, as shown in Fig. \ref{HSfunction}. The HOG visualization, extracted from the top layer, highlights the direction of intensity changes, providing a clear representation of texture and structural features. The H\&S histograms, derived from the bottom layer, illustrate the hue and saturation distributions of the images. Additionally, the global HOG histogram summarizes feature values across nine directions. 

Image I-1, which shares a similar color profile with the template I-T, is effectively distinguished using HOG visualizations derived from top-layer features. Image I-2, despite having a comparable HOG structure, is differentiated through the H\&S histogram. Image I-3, which exhibits similarities in both color and structural features, is identified by analyzing the variance of global HOG features extracted from the bottom layer. The results indicate that oil palm tree leaflets exhibit greater variation than other palm species. Top-layer HOG features, extracted at a low resolution, efficiently capture general structural patterns but may overlook fine details needed for species differentiation. In contrast, bottom-layer HOG features, obtained at a higher resolution, preserve finer details, and their variance enhances species distinction. However, this feature extraction step relies on high-resolution images for optimal performance.

A key limitation of the detection method arises in densely clustered trees with high canopy overlap, as demonstrated in Image I-F. Significant canopy overlap alters the HOG feature visualization compared to Image I-T, thereby reducing detection accuracy. The detection process relies on an SVM model trained to classify HOG features. Performance evaluation on the test set shows that the model achieved an accuracy of 92.91\%, a precision of 90.37\%, and a recall of 96.06\%. These results show that the proposed method provides sufficient accuracy for oil palm tree detection.

\begin{figure}[!t]
    \centering
    \includegraphics[width=2.8in]{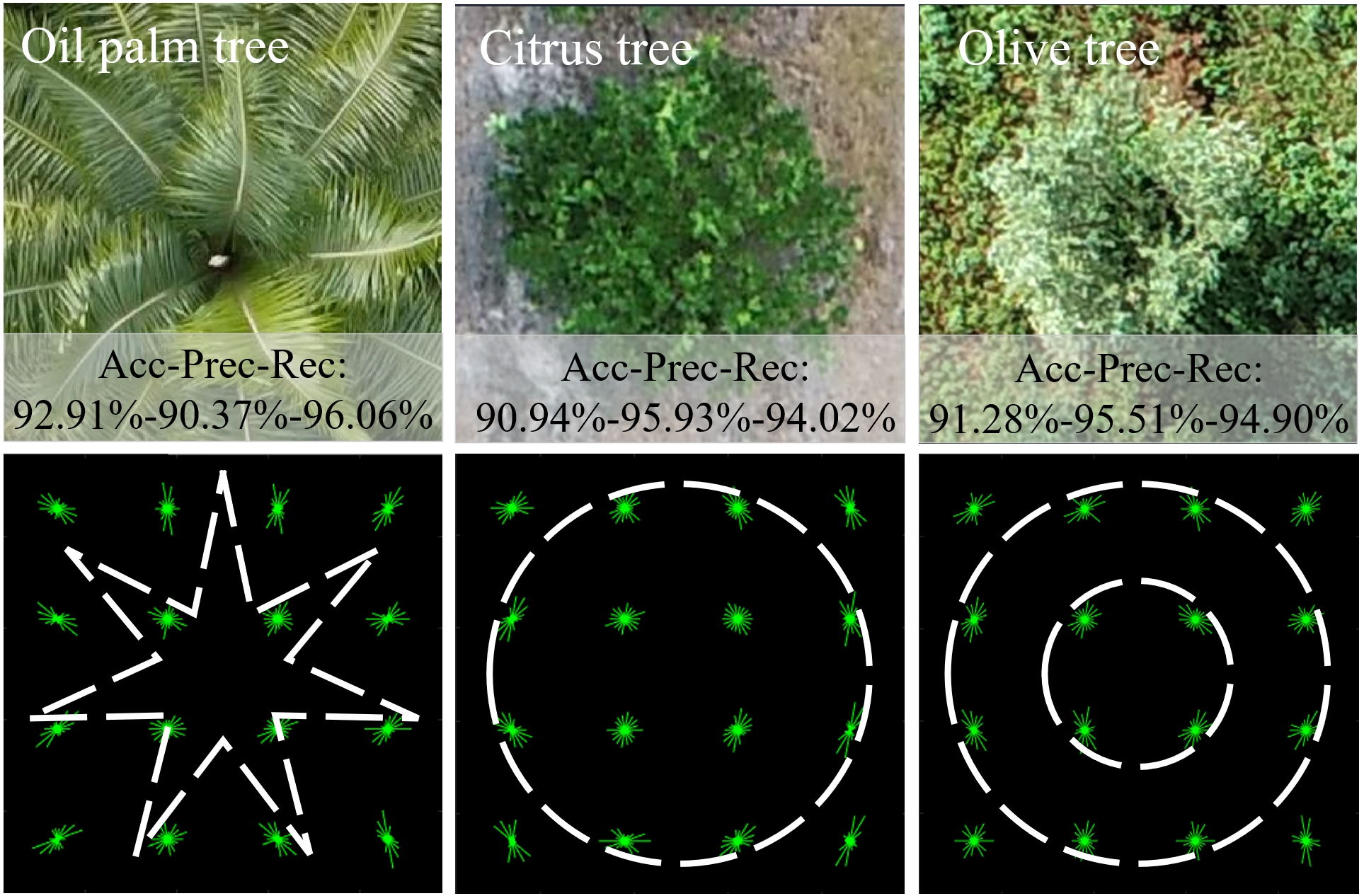}
    \caption{HOG feature visualization and SVM classification performance for different tree types, evaluated in terms of Accuracy (Acc), Precision (Prec), and Recall (Rec).}
    \label{Multipletree}
\end{figure}

To assess the generalization of our detection method, it was applied to citrus and olive trees. Fig. \ref{Multipletree} presents their HOG features alongside the performance evaluation of the SVM classifier. The overall structure of HOG features is outlined with white dashed lines, revealing distinct patterns for each species: a star-like shape for oil palm, a single-circle shape for citrus, and a double-circle shape for olive. The high accuracy, precision, and recall of the SVM classifier demonstrate the robustness of our method across different tree species.

Table 1 summarizes real-time performance metrics, including FPS, CPU usage, and temperature, comparing our method with lightweight deep learning models. The evaluation was conducted on an onboard processing unit equipped with an Allwinner A33 chipset and an integrated camera. The entire system measures $1.7~in^3$ and weighs $17~g$. The tests were designed to evaluate computational efficiency under real-world conditions. The results indicate that while MobileNetV2-SSD and EfficientDet achieve perfect recall, their lower accuracy and precision suggest a higher rate of false positives. 
Moreover, our method demonstrates superior efficiency, operating with lower CPU usage and temperature, leading to a higher FPS. In contrast, MobileNetV2-SSD and EfficientDet are more resource-intensive, resulting in significantly lower FPS. These findings highlight the advantage of our method in real-time embedded applications, offering efficient performance without compromising detection accuracy.

\begin{table}
\begin{center}
\renewcommand{\arraystretch}{1.5} 
\caption{Comparison of our method with lightweight deep learning models in terms of detection performance and resource usage on embedded devices.}
\label{tab1}
\begin{tabular}{ m{1.5cm}  m{2.5cm}  c  c }
\toprule
Methods & Acc - Prec - Rec(\%) & \makecell{CPU \\ usage(\%) - temp(°C)} & FPS \\
\midrule
Our methods & 92.91 - 90.37 - 96.06 & 114.8 - 44 & 8.9 \\

MobileNetV2-SSD & 77.87 - 69.61 - 100.00 & 326.8 - 68 & 5.8 \\

EfficientDet & 85.77 - 78.26 - 100.00 & 298.1 - 71 & 1.9 \\
\bottomrule
\end{tabular}
\end{center}
\end{table}

Beyond tree detection, we evaluated the overall navigation performance through multiple flight tests. In Map 1, the three test trajectories (Fig. \ref{FlightResults1}) demonstrate that the MAV successfully detected and navigated over Trees 1 to 6 in sequence, dynamically updating its flight path. To further assess adaptability, additional tests were performed on Map 2, where trees were not aligned in a straight column (Fig. \ref{FlightResults2}, left). The trajectory plot illustrates that the MAV correctly identified and maneuvered toward all seven trees, demonstrating its ability to navigate irregular layouts.

Building upon these evaluations, Tests 7–9 were conducted in a more complex setting on Map3, where trees were irregularly distributed across two planting columns (Fig. \ref{FlightResults2}, right). Trees 3–5 formed one cluster, while Trees 6–8 formed another. Despite the increased complexity, the MAV effectively distinguished individual trees, adjusted its trajectory accordingly, and maintained accurate navigation.

These results also highlight the robustness of our navigation strategy in handling real-world challenges:
\subsubsection{Bypassing non-target trees}

In Test 2, while approaching Tree 3, the MAV encountered a non-target tree outside the planting column. Instead of deviating, it identified the tree’s position but remained focused on its intended trajectory, as reflected in the onboard detection results indicated in a yellow box (Fig. \ref{FlightResults2}, left). This selective targeting ensures precise navigation within designated plantation areas.

\subsubsection{Recovery mechanism}

Test 3 demonstrated this capability when the MAV lost sight of Tree 6, leading to a slight deviation. Similarly, in Test 6, Tree 5 was temporarily lost, but the MAV quickly realigned itself using stored position data and resumed its trajectory. A zoomed-in trajectory plot (Fig. \ref{FlightResults2}, right) highlights these corrective maneuvers. Even in dense clusters, such as in Test 8, the recovery mechanism enabled the MAV to accurately reach Tree 5 despite canopy overlap.

\begin{figure}[!t]
    \centering
    \includegraphics[width=0.45\textwidth]{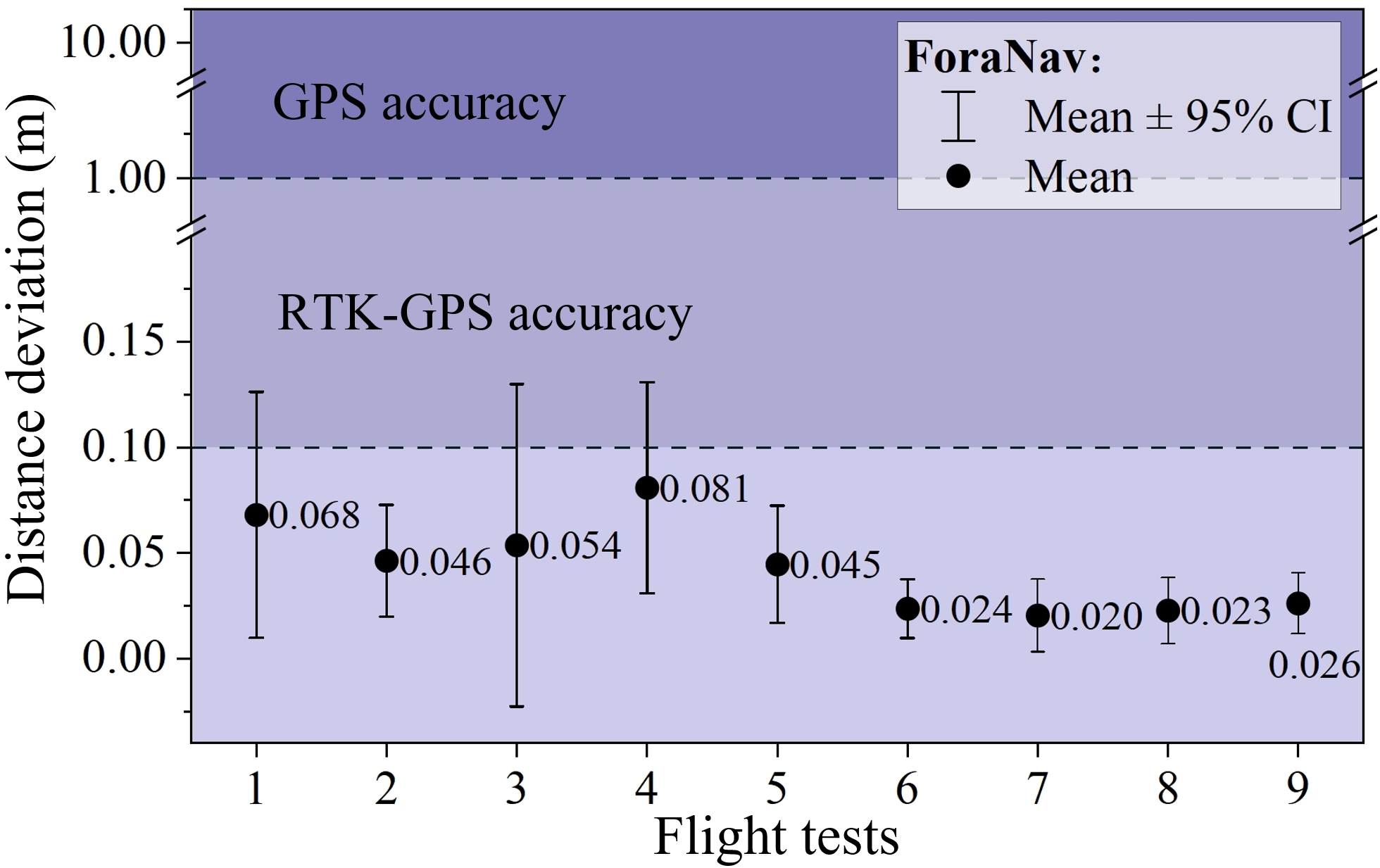}
    \caption{Distance deviation of flight trajectories from target trees in Test 1 to Test 9 using ForaNav. The mean represents the average distance deviation in each test. The mean ± 95\% CI (Confidence Interval) indicates the range within which the true mean is expected to lie within 95\% confidence.}
    \label{Quantitydata}
\end{figure}

To quantify accuracy, Fig. \ref{Quantitydata} presents the distance deviation of the MAV from the actual tree positions across all tests, using ForaNav. Compared to conventional GPS and RTK-GPS benchmarks \cite{yin2022modeling}, our approach achieved a mean deviation below 0.1 m, demonstrating superior precision. Crucially, this level of accuracy was achieved without reliance on predefined tree coordinates, reinforcing the potential for fully autonomous navigation in plantation environments.

\section{Conclusions}
\label{sec:conclusion}

This work presented ForaNav, an insect-inspired online target-oriented navigation method designed for tree plantation management tasks. Our approach integrates a real-time tree detection algorithm using an enhanced hierarchical HOG method and a navigation strategy inspired by insect foraging behavior, incorporating a robust recovery mechanism. The detection method, leveraging H\&S similarity comparison and hierarchical HOG feature extraction, effectively distinguished oil palm trees from the background and other visually similar palm species. Moreover, it generalized well to different tree types. Compared to lightweight deep learning models, our method achieved lower CPU usage and temperature while maintaining a higher FPS, making it more efficient for real-time operation. Experimental results demonstrated that the MAV successfully detected and approached all trees across various scenarios without prior tree information. This work lays the foundation for fully autonomous plantation tasks, such as precision spraying and fruit ripeness detection, further advancing plantation management in precision agriculture. The demonstration videos are
accessible\footnote{Video:\url{ https://www.youtube.com/playlist?list=PLMFhxEZKPi2r-FzVmvgXQk0VUCKnhJUMF}}, and the project code, datasets, and trained models are available in our repository\footnote{Repository:\url{ https://github.com/iAerialRobo/Online-Target-oriented-Navigation-for-Micro-Air-Vehicles-in-Tree-Plantations.git}}.


\bibliographystyle{IEEEtran}
\bibliography{IEEEabrv, Reference.bib}
\end{document}